\documentclass[10pt, a4paper]{article}

\usepackage[final]{lrec-coling2024} 
\usepackage{cite}
\usepackage{natbib}
\usepackage{threeparttable}

\title{\textbf{Probing Large Language Models in Reasoning and Translating Complex Linguistic Puzzles}}
\name{Zheng-Lin Lin, Yu-Fei Shih, Shu-Kai Hsieh} 
\address{\textbf{National Taiwan University} \\
         b09208026@ntu.edu.tw, yfshih@nlg.csie.ntu.edu.tw, shukaihsieh@ntu.edu.tw\\}

\abstract{
This paper investigates the utilization of Large Language Models (LLMs) for solving complex linguistic puzzles, a domain requiring advanced reasoning and adept translation capabilities akin to human cognitive processes. We explore specific prompting techniques designed to enhance LLMs' ability to reason and elucidate their decision-making pathways, with a focus on Input-Output Prompting (IO), Chain-of-Thought Prompting (CoT), and Solo Performance Prompting (SPP). Utilizing datasets from the Puzzling Machine Competition and various Linguistics Olympiads, we employ a comprehensive set of metrics to assess the performance of GPT-4 0603, a prominent LLM, across these prompting methods. Our findings illuminate the potential of LLMs in linguistic reasoning and complex translation tasks, highlighting their capabilities and identifying limitations in the context of linguistic puzzles. This research contributes significantly to the broader field of Natural Language Processing (NLP) by providing insights into the optimization of LLM applications for improved reasoning and translation accuracy, thereby enriching the ongoing dialogue in NLP advancements.
 \\ \newline \Keywords{Machine Translation, Machine Reasoning, Large Language Models, Explainability, Linguistic Olympiad, Text Analytics} 
}

\begin{document}

\maketitleabstract

\begin{figure*}[htp]
    \centering
    \includegraphics[width=16cm]{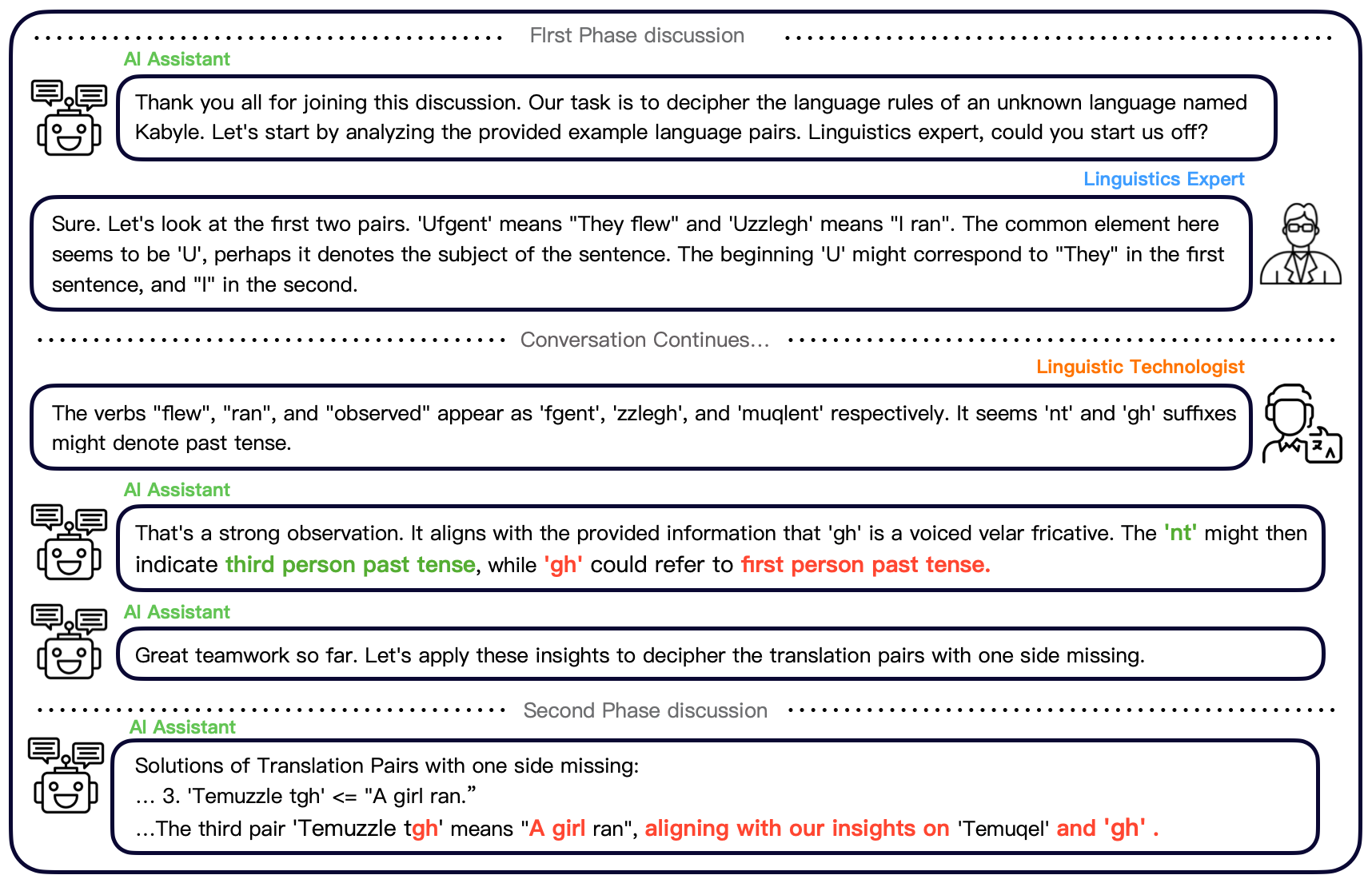}
    \caption{Illustration of Rule Contradiction in language Kabyle: GPT-4 breaches its own established rule, wherein 'gh' is designated to signify the first person past tense.}

    \label{spp_rule}
\end{figure*}

\begin{figure*}[htp]
    \centering
    \includegraphics[width=16cm]{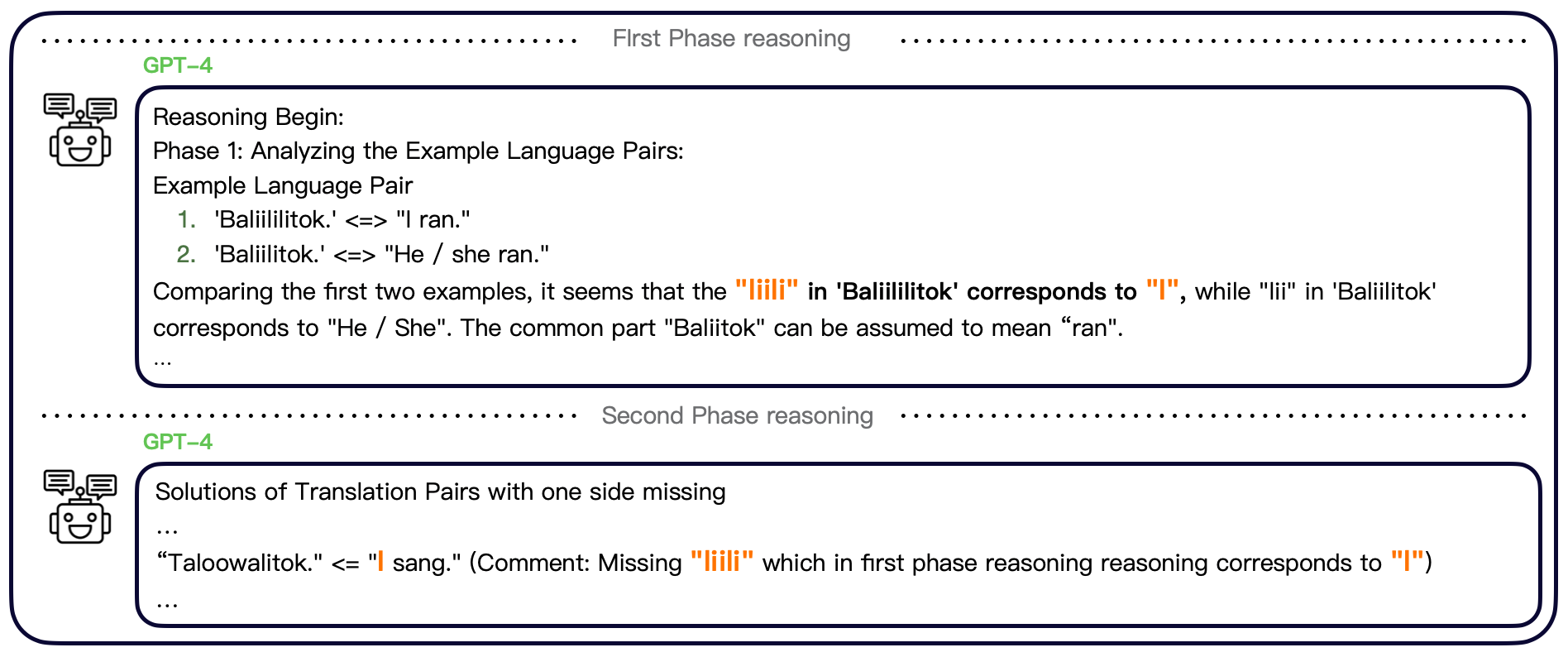}
    \caption{An example of dictionary contradiction within GPT-4's reasoning process using CoT Prompting on the Rosetta Stone Problem of Choctaw. }

    \label{cot_dic}
\end{figure*}

\section{Introduction}
The exploration of human cognitive systems has unveiled a fascinating dichotomy: System 1, responsible for quick, intuitive reactions, and System 2, which governs our capacity for complex reasoning, thereby operating at a slower pace and demanding more energy \citep{tfs}. This bifurcation in cognitive processes is starkly evident in the realm of language translation. While the task of translating a single word may lean heavily on the rapid, reflexive capabilities of System 1, akin to a simple lookup in a dictionary, the translation of entire sentences plunges into the realm of System 2, requiring a deeper analytical engagement. This engagement is especially crucial in contexts where references are scant and the linguistic puzzle complex.

Within this framework, the Rosetta Stone Problem emerges as a quintessential challenge, embodying the essence of System 2 reasoning within linguistic puzzles. Featured prominently in the International Linguistics Olympiad~\footnote{https://ioling.org}, this problem demands from its solvers not only the translation of texts between two languages with limited references but also the construction and application of a mini-grammar and vocabulary deduced from the given material \citep{Rosetta_IOL}. Such tasks underscore the profound complexity and the intricate cognitive engagement required, mirroring the deliberative, analytical processes characteristic of System 2 thought.

The advent of Large Language Models (LLMs) has opened new vistas in addressing complex linguistic challenges, such as the Rosetta Stone Problem. This paper delves into the potential of LLMs to navigate these intricacies, with a particular focus on specific prompting techniques believed to significantly enhance the models’ ability to translate with greater accuracy and elucidate their reasoning paths, paralleling human cognitive processes. By exploring the efficacy of these prompting strategies, we aim to illuminate the nuances of LLM reasoning, their potential for error detection and correction, and the implications for complex translation tasks. Through our examination, we seek not only to advance our understanding of LLM capabilities in complex linguistic reasoning and translation but also to contribute to the broader field of Natural Language Processing (NLP) by fostering innovative developments and optimizing LLM applications, particularly in linguistic reasoning and machine translation, thus enriching the ongoing dialogue in NLP advancements.

\section{Background}
The Puzzling Machine Challenge~\footnote{https://ukplab.github.io/PuzzLing-Machines/} has been a significant landmark in showcasing the capabilities of LLMs, particularly with the application of methodology by \citet{PuzzGPT} that utilized ChatGPT in conjunction with Input Output Prompting (IO) to achieve unprecedented success in solving Rosetta Stone Problems \citep{Puzz_Machine}. This success highlights the potential of LLMs to outperform traditional methods in linguistic puzzles, setting a new benchmark for future research and application.

Translating within the constraints of Rosetta Stone problems, however, introduces substantial challenges. The structures of languages often diverge significantly, a fact that can lead to inaccuracies in the mini-grammars and dictionaries derived by solvers. This discrepancy poses a notable challenge, emphasizing the necessity for LLMs to not only generate translations but also to navigate and correct errors in their initial linguistic assumptions \citep{Rosetta_IOL}.

Recent developments in prompting techniques have brought forth a variety of methods aimed at eliciting more sophisticated reasoning from LLMs. Among these, Chain-of-Thought (CoT), Self-Consistency (SC-CoT), Tree-of-Thought (ToT), and Solo Performance Prompting (SPP) stand out for their potential to facilitate System 2-like cognitive processes in machines \citep{CoT, SC, ToT, SPP}. These techniques represent a significant stride toward enhancing the depth of reasoning LLMs can exhibit, providing a new lens through which the complexities of linguistic puzzles can be approached.

\section{Experiments}
In this section, we delineate our experimental setup designed to evaluate the reasoning capabilities of Large Language Models (LLMs) in addressing Rosetta Stone Problems. Our investigation centers on the application of GPT-4 0603 (hereafter GPT-4), a specific iteration of the GPT-4 model, to explore its performance across a carefully curated dataset using a variety of prompting methods. 

\subsection{Dataset}
We utilized two datasets for our evaluation: one from the Puzzling Machine Challenge, as detailed by \citet{Puzz_Machine}, and another compiled from the United Kingdom Linguistics Olympiad (UKLO)\footnote{https://www.uklo.org} and the North American Computational Linguistics Open Competition (NACLO)\footnote{https://naclo.org}. The Puzzling Machine Challenge dataset, featuring around 100 problems in 81 languages \citep{Puzz_Machine}, was our primary source, from which we selected 86 unlabeled problems to test GPT-4's accuracy. 

\begin{figure}[htp]
    \centering
    \includegraphics[width=7cm]{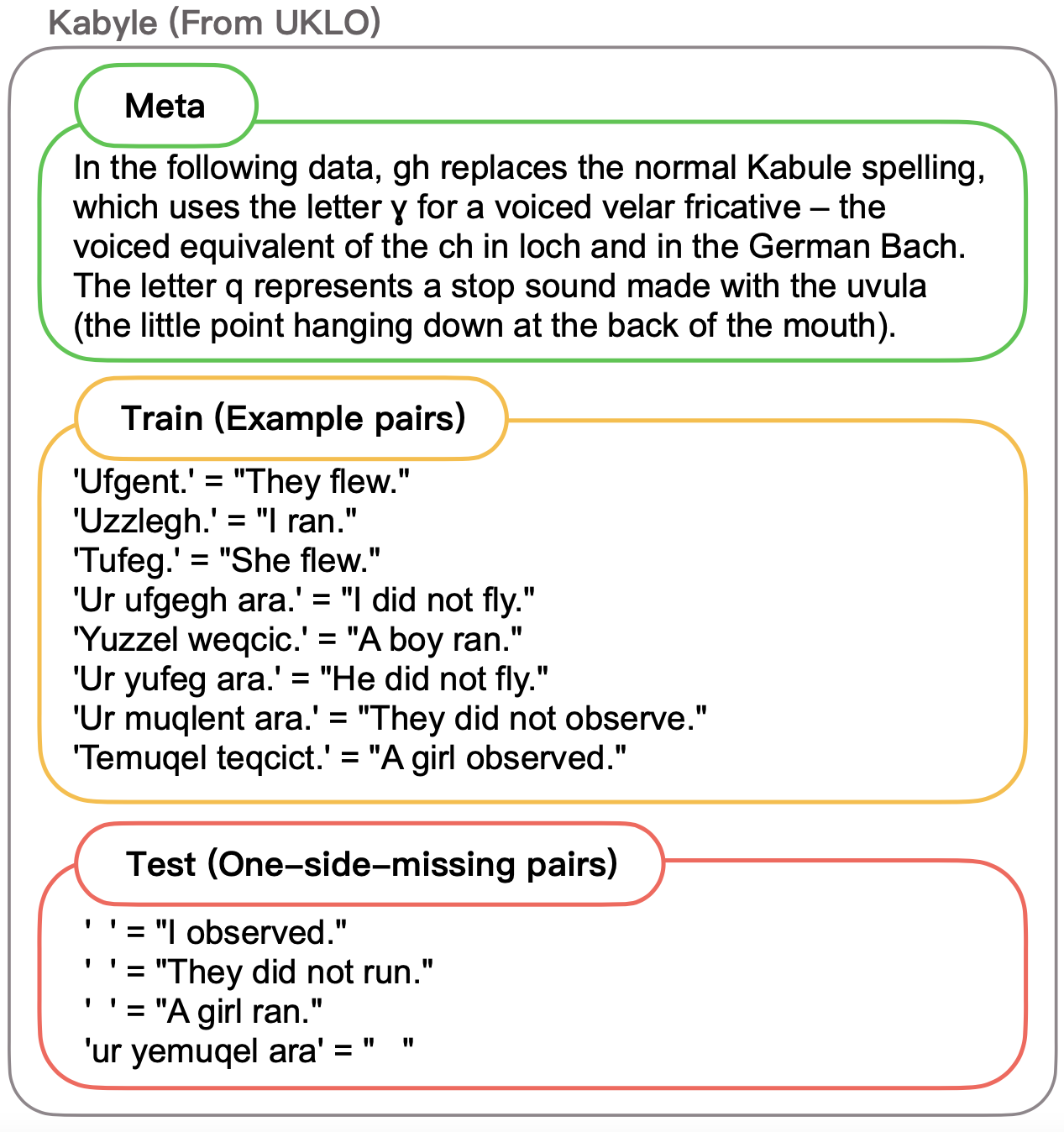}
    \caption{Problem format example of northern Algeria language, Kabyle, collected and refined from UKLO.}
    \label{pm_ex}
\end{figure}

Our second dataset consists of 28 problems either directly obtained or adapted into Rosetta Stone puzzles from LO competitions. These problems were modified to match the format used in the Puzzling Machine Challenge, ensuring consistency in evaluation. Each problem is structured into a Meta section, providing crucial information on the foreign language, a Train Set of translation pairs for deriving rules, and a Test Set where one side of the pair is missing. An illustrative example of this format is shown in Figure~\ref{pm_ex}. 

\subsection{Prompting Methods}

\textbf{Input-Output Prompting (IO) and Zero-Example Prompting (ZeroEx):}
IO prompting, as used by \citet{PuzzGPT} with ChatGPT, introduces the task without detailed solving instructions, aiming for GPT-4 to generate answers without reasoning paths. ZeroEx Prompting, a variant of IO we modified which excludes example pairs to test if GPT-4 recognizes the language, focusing on raw answer generation. 

\textbf{Two-Phase reasoning strategy.} Tailored to the unique challenges of Rosetta Stone Problems, this strategy directs GPT-4 through two reasoning phases. Initially, the model analyzes rules and vocabularies from examples, followed by rule application and potential revision in the test phase. This setup ensures comprehensive discussion of all language pairs and necessitates explanations for assumptions, facilitating an in-depth analysis of GPT-4’s reasoning.

\textbf{Chain-of-Thought Prompting:}
Enhancing the IO method, CoT introduces a step-by-step reasoning instruction, based on findings by \citet{CoT} that such an approach improves LLM performance on reasoning-intensive tasks. We integrated CoT with the two-phase strategy, incorporating organized reasoning directives.

\textbf{Multi-Experts Self-Collaboration:} 
Drawing from Solo Performance Prompting (SPP) by \citet{SPP}, which simulates discussions among various personas, this method assigns GPT-4 as a facilitator to enhance engagement. Proven effective in pretests, it's adapted here to align with the two-phase reasoning approach, aiming to extract GPT-4’s internal knowledge while maintaining its reasoning capabilities.

\section{Evaluation}
This section outlines the methodology adopted to evaluate GPT-4's proficiency in solving Rosetta Stone Problems, utilizing two distinct datasets: the Puzzling Machine Competition data and a dataset compiled from various Linguistic Olympiads (LO). Central to our investigation are two primary objectives: firstly, to assess the impact of various prompting techniques on GPT-4's ability to generate reasoning paths that parallel human cognitive processes, and secondly, to illuminate the nuances of LLM reasoning, including its potential for error detection and correction. Given the Puzzling Machine provides an online judging system and the answers to all problems are not publicly accessible, our analysis will not examine the reasoning path of GPT-4 on this dataset. 

We leveraged the evaluation tool made available on their official website\footnote{\url{https://eval.ai/web/challenges/challenge-page/2150/overview}}, as recommended by \citet{Puzz_Machine}. This approach ensures our assessment aligns with the competition's established metrics, facilitating a standardized evaluation of GPT-4's performance without delving into the reasoning paths due to the aforementioned constraints.

\textbf{For translations from English to the Unknown Language:}
\begin{itemize}
    \item \textbf{BLEU-2:} A word-level based metric that assesses the quality of the generated text by comparing it with reference texts, using bigrams to provide a balance between precision and recall \citep{bleu}.
    \item \textbf{characTER:} A character-level metric designed to evaluate translation accuracy by considering edits at the character level, which is particularly useful for capturing finer linguistic nuances \citep{characTer}.
    \item \textbf{chrF:} Another character-level metric that calculates F-scores based on character n-grams, facilitating a detailed assessment of translation quality \citep{chrf}.
\end{itemize}

\textbf{For translations from the Unknown Language to English:}
We utilized embeddings generated by the \textit{all-MiniLM-L6-v2} model from Sentence Transformers \citep{CosSim} for the English sentences. The evaluation score is derived by calculating the cosine similarity (CosSim) between the embedding vectors of the generated and reference texts, offering a measure of semantic similarity.

\textbf{Overall Performance Evaluation:}
\begin{itemize}
    \item \textbf{Exact Match (EM):} This metric assigns a score of 1 if GPT-4's generated expression exactly matches the reference expression, and 0 otherwise. EM is used to evaluate the model's performance across both translation directions, providing a direct measure of accuracy.
\end{itemize}

By employing these metrics, our evaluation framework aims to comprehensively assess the capabilities of GPT-4 in solving Rosetta Stone Problems, taking into account both the precision of translation and the semantic accuracy of the generated text.




\section{Result}
In this section, we delve into the performance evaluation of GPT-4 regarding its ability to solve Rosetta Stone Problems across varying datasets and experimental conditions.
\subsection{Performance Analysis Using the Linguistics Olympiad Dataset}
The performance of different methodologies in the Linguistics Olympiad dataset is quantitatively assessed through Table~\ref{res_LO_que}, which presents the average scores computed by dividing the total scores for each query by the number of queries. This analysis reveals that the Information Ordering (IO) scores consistently surpass those achieved by other methods, while the ZeroEx approach lags behind in all metrics. A comparative evaluation of the Self-Paced Learning (SPP) and Chain of Thought (CoT) methods shows similar overall performance, with CoT yielding higher Exact Match (EM) scores and better performance in metrics related to the translation from English to unknown languages (CharacTER, ChF-3, and BLEU-2). Conversely, SPP performs better in the CosSim metric, indicating a superior ability to translate from unknown languages to English.

To mitigate potential biases introduced by languages with a higher query count in the LO dataset, we refer to Table~\ref{res_LO_pro}. This table adjusts for language representation by first averaging the scores within each language and then averaging these across all languages. The findings align with those in Table~\ref{res_LO_que}, highlighting the superior performance of IO and the comparatively lower scores of ZeroEx. CoT demonstrates an advantage in EM and English-to-unknown language translation metrics over SPP, while both methods show comparable performance in the CosSim score.

\subsection{Analysis Based on the Puzzling Machine Competition Dataset}

Table~\ref{PuzzRes} details the performance metrics for SPP, CoT, and IO using data from the Puzzling Machine Competition. Despite differences in the metrics compared to those used for the LO dataset analysis, they convey analogous insights. Metrics prefixed with "FE" are designed to evaluate translations from unknown languages to English, whereas "EF" metrics assess translations from English to unknown languages. According to Table~\ref{PuzzRes}, IO outperforms both SPP and CoT across all metrics. Interestingly, the dataset reveals a reversal in the comparative performance of SPP and CoT observed in the LO dataset analysis: SPP marginally outperforms CoT in EM and English-to-unknown language translation metrics, while CoT exhibits superior performance in translating from unknown languages to English.

\begin{table*}
\caption{Puzzling Machine Competition Result}
\begin{threeparttable}
\begin{tabular}{lrrrrrrr}
\label{PuzzRes}
 prompting method   &    EM &   FE\_CTER\tnote{a} &   FE\_CHRF &   FE\_BLEU &   EF\_CTER\tnote{b} &   EF\_CHRF &   EF\_BLEU \\
\hline
 SPP               & 31.83 &     66.92 &     70.99 &     56.91 &     69.74 &     72.13 &     39.83 \\
 CoT               & 31.37 &     70.64 &     73.78 &     60.19 &     68.64 &     71.25 &     38.39 \\
 IO                & 33.79 &     73.14 &     75.79 &     61.37 &     74.86 &     75.56 &     42.65 \\
\hline
\end{tabular}
\begin{tablenotes}
    \item[a] FE denotes translation from an unknown language to English
    \item[b] EF denotes translation from English to an unknown language
\end{tablenotes}
\end{threeparttable}
\end{table*}
\subsection{Analysis of GPT-4's Linguistic Proficiency Using the Linguistic Olympics Dataset}

\begin{figure*}[h!]
    \centering
    \includegraphics[width=1.0\textwidth]{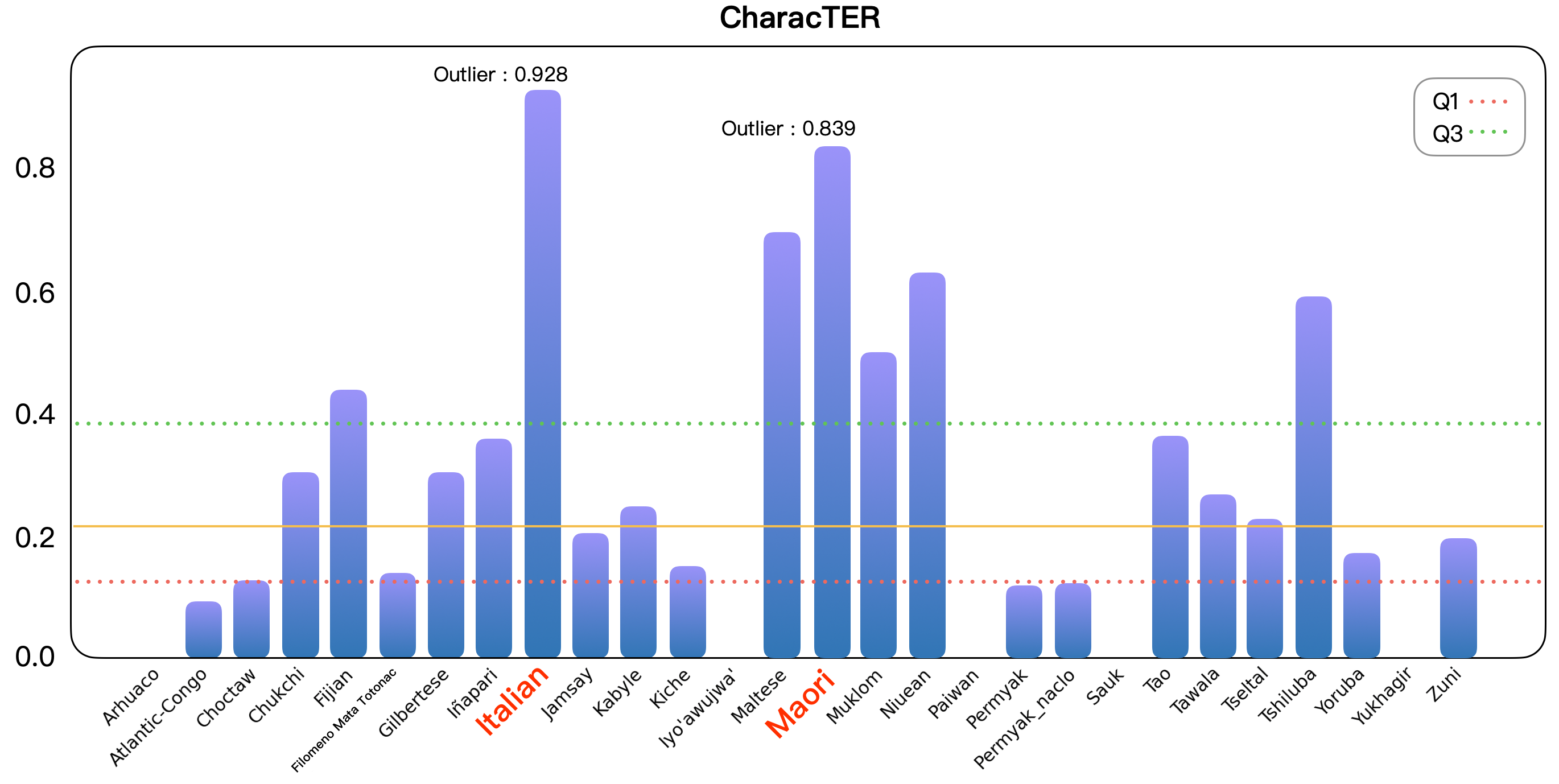}
    \caption{characTER score of zero example}
    \label{fig:example2}
\end{figure*}

In this study, we evaluated GPT-4's linguistic capabilities across various languages represented in the Linguistic Olympics Dataset, utilizing the ZeroEx metric as a primary analytical tool. Our analysis, grounded in five distinct metrics, revealed a consistent distribution pattern across all languages assessed. Consequently, we chose to highlight the CharacTER scores of the ZeroEx method for all 28 languages in Figure~\ref{fig:example2}, as a representative illustration of the overarching trends observed with other metrics. This graphical representation allows for categorization of languages into three distinct tiers based on GPT-4's proficiency:

\begin{itemize}
\item \textbf{Limited Proficiency Languages:} This category includes Arhuaco, Iyo'awujwa, Paiwan, Sauk, and Yukhagir. For these languages, GPT-4 exhibited a complete lack of translation capability, indicating minimal to no knowledge.
\item \textbf{High Proficiency Languages:} Italian and Maori are placed within this tier. GPT-4 demonstrated a comprehensive understanding of these languages, accurately fulfilling translation requests without significant errors.
\item \textbf{Moderate Proficiency Languages:} The remainder of the languages fall into this intermediate category. While GPT-4 is capable of generating translations for these languages, the accuracy and correctness of the output vary, indicating a level of proficiency that falls between the two extremes outlined above.
\end{itemize}

\begin{table*}
\begin{center}
\caption{Average Performance of all Queries}
\begin{tabular}{lrrrrr}
\label{res_LO_que}
 Prompting Method   &    EM &   CosSim &   CharacTER &   ChF-3 &   BLEU-2 \\
\hline
 SPP                & 0.171 &    0.731 &       0.516 &   0.539 &    0.29  \\
 CoT                & 0.183 &    0.704 &       0.618 &   0.643 &    0.351 \\
 IO                 & 0.217 &    0.767 &       0.642 &   0.67  &    0.384 \\
 ZeroEx               & 0.071 &    0.344 &       0.317 &   0.342 &    0.19  \\
\hline
\end{tabular}
\end{center}
\end{table*}

\begin{table*}
\begin{center}
\caption{Average Performance of all Languages}
\begin{tabular}{lrrrrr}
\label{res_LO_pro}
 Prompting Method   &    EM &   CosSim &   CharacTER &   ChF-3 &   BLEU-2 \\
\hline
 SPP                & 0.077 &    0.734 &       0.509 &   0.546 &    0.249 \\
 CoT                & 0.085 &    0.734 &       0.603 &   0.633 &    0.296 \\
 IO                 & 0.103 &    0.765 &       0.626 &   0.664 &    0.334 \\
 ZeroEx               & 0.025 &    0.318 &       0.288 &   0.315 &    0.142 \\
\hline
\end{tabular}
\end{center}
\end{table*}

\section{Discussion}
The results from our experiment presented a surprising observation: IO consistently outperformed CoT and SPP on all metrics assessed. Before providing the potential reason for this deviation, it's pertinent to discuss the challenges that reduce the performance of CoT and SPP.

The strategy in solving Rosetta Stone Problems entails recognizing underlying linguistic rules and then creating a dictionary to bridge English with the unknown language. This demands iterative refinement as the example language pairs are cross-referenced.

\begin{figure}[htp]
    \centering
    \includegraphics[width=7.5cm]{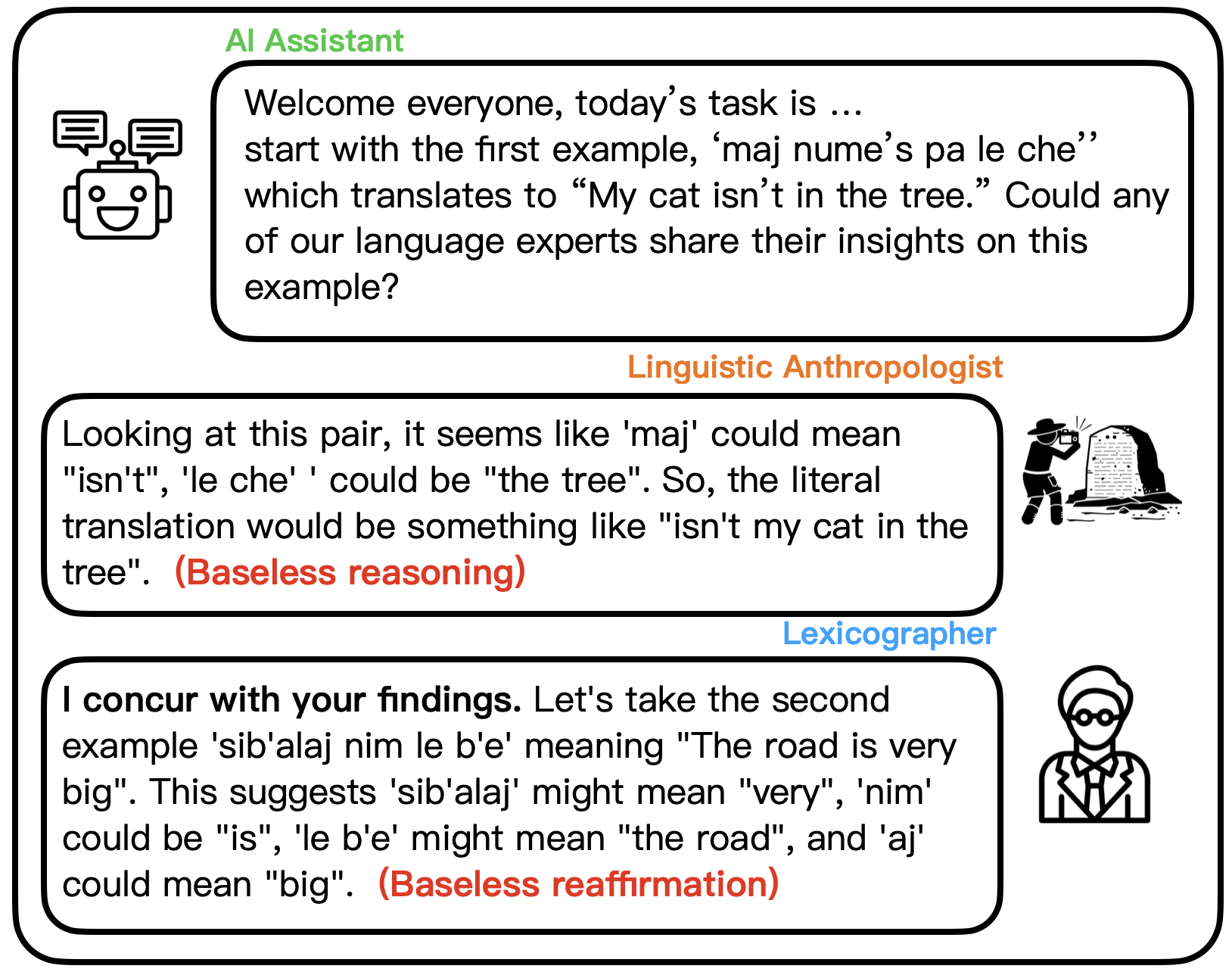}
    \caption{The figure shows the baseless assumption occurring in the CoT discussion on Kiche language. Linguistic anthropologists first propose baseless vocabulary pairs and lexicographers reaffirm the opinion.}
    \label{A1}
\end{figure}

In the rule identification process, CoT frequently makes assumptions about linguistic rules and potential dictionary pairings without supplying adequate justification. As this process iteratively go through the following example language pairs, there is almost no retrospective corrections to these assumptions. Instead, CoT often reaffirms the pre-established rules and dictionary pairings as correct. See Figure~\ref{A1} for the example.

Additionally, CoT's reasoning process revealed a bias. When confronted with linguistic challenges unfamiliar to GPT-4, CoT regularly relied on implicit grammar rules stemming from English. In contrast to the familiar language of GPT-4 like Italian, CoT clearly explained the derivation and application of language rules that determine word choices based on grammatical gender. However, the reasoning process was not comprehensive enough to provide the complete solution to the puzzle.

Challenges also emerge with the SPP method. Conversations between expert personas seldom added depth to the discussion. Out of 28 language puzzles, expert personas challenged each other's assumptions in just one instance. The dialogue tends to remain overly harmonious, whereas real-life interactions among diverse roles often elicit broader perspectives.

A closer look at the second-phase GPT-4's translation capabilities revealed another limitation. The model frequently outputs translations without elucidating the reasoning behind these choices. Furthermore, because of the incomplete rules from the first phase, GPT-4 always applied rules or vocabulary not discussed in the first phase. It also occurred when translating words that varied from those in the pre-established dictionary, both CoT and SPP provide alternate translations without justification.

A pivotal observation from our study involves the contradictory nature of GPT-4's reasoning process. Specifically, we have identified instances where the answers provided by GPT-4 were at odds with the reasoning framework it initially established. We have classified these contradictions into two categories: dictionary contradictions and rule contradictions. This classification delineates instances where GPT-4's conclusions either contravene the lexical parameters (dictionary contradictions) or the logical premises (rule contradictions) it had previously set. See Figure~\ref{cot_dic} for examples of dictionary contradictions and Figure~\ref{spp_rule} for rule contradictions encountered in our analysis.  This suggests GPT-4's generated reasoning might not genuinely mirror its internal thought process when tackling the Rosetta Stone problems, leading to varying outcomes. However, it's essential to highlight that the identified contradictions in the second phase don't necessarily correlate with output quality. There were instances where GPT-4 disregarded incorrect pre-established rules, resulting in the correct answer, as well as cases where it overlooked the correct rules, producing an erroneous response. The relationship between these inconsistencies and performance outcomes warrants further exploration.

In summary, GPT-4's challenges in addressing the Rosetta Stone Problems primarily revolve around its inability to provide a thorough reasoning pathway and accurate translation pairs consistently. This leads us to the assumption that the superior performance of the IO method could stem from GPT-4's incomplete rules and dictionary causing distractions or adding noise to the translation process. In contrast, the IO method, by relying on example language pairs only, offers answers without the convoluted reasoning process.

\section{Conclusion}

In light of our study, which delved into GPT-4’s reasoning capabilities concerning linguistic puzzles, several limitations emerged, emphasizing the need for continued exploration in this area.

We conducted a detailed analysis of GPT-4’s reasoning process primarily using the LO dataset, which comprises only 28 language puzzles. Given the limited size of this dataset, there’s a potential for bias on the puzzles we selected. Future studies should consider a more expansive and diverse dataset to ensure comprehensive insights.

Furthermore, our results(see figure) indicate that GPT-4 may have varying levels of familiarity with different languages. Throughout our investigation, we employed both SPP and CoT prompting approaches for the entire dataset. It’s important to recognize that identifying various familiarity levels for GPT-4 and designing separate experiments for each level might yield more detailed insights into GPT-4’s reasoning process.

Our findings highlight challenges in GPT-4’s ability to generate linguistic rules and dictionaries and subsequently apply them to unknown language translations. An avenue for future research could explore alternative prompting methods that not only guide GPT-4 to produce accurate answers but also elucidate the comprehensive reasoning processes leading to those conclusions from the given example language pairs only.

Our analysis shows that the IO method consistently surpassed both SPP and CoT in performance. It remains crucial to delve deeper into the reasons behind the superior efficacy of the direct IO method. Moreover, a comprehensive examination is required to assess our assumption that the incomplete or errors in the reasoning processes of GPT-4’s response could impact its output quality and performance.

\nocite{*}
\section{References}\label{sec:reference}

\bibliographystyle{lrec-coling2024-natbib}
\bibliography{lrec-coling2024-example}


\end{document}